\documentclass{article}
\usepackage{ijcai20}

\usepackage{times}
\usepackage{soul}
\usepackage{url}
\usepackage[hidelinks]{hyperref}
\usepackage[utf8]{inputenc}
\usepackage[small]{caption}
\usepackage{graphicx}
\usepackage{amsmath}
\usepackage{amsthm}
\usepackage{amssymb}
\usepackage{booktabs}
\usepackage{algorithm}
\usepackage{algorithmic}
\usepackage[dvipsnames]{xcolor}
\usepackage{float}
\definecolor{myred}{rgb}{0.8,0,0}
\newcommand{\todo}[1]{}
\renewcommand{\todo}[1]{{\color{myred} Todo: {#1}}}

\newcommand{\mycite}[1]{\citeauthor{#1}~\shortcite{#1}}

\usepackage{scalefnt}
\usepackage{enumitem}

\usepackage{makecell}

\title{Automatic Curriculum Learning For Deep RL: A Short Survey}

\author{
R\'emy Portelas$^1$
\and
C\'edric Colas$^1$\and
Lilian Weng$^2$\and
Katja Hofmann$^3$\And
Pierre-Yves Oudeyer$^1$
\affiliations
$^1$Inria, France\\
$^2$OpenAI, USA\\
$^3$Microsoft Research, UK\\
\vspace{0.12cm}
remy.portelas@inria.fr}

\begin{document}

\maketitle

\begin{abstract}
    Automatic Curriculum Learning (ACL) has become a cornerstone of recent successes in Deep Reinforcement Learning (DRL). 
    These methods shape the learning trajectories of agents by challenging them with tasks adapted to their capacities.
    In recent years, they have been used to improve sample efficiency and asymptotic performance, to organize exploration, to encourage generalization or to solve sparse reward problems, among others.
    To do so, ACL mechanisms can act on many aspects of learning problems. 
    They can optimize domain randomization for Sim2Real transfer, organize task presentations in multi-task robotic settings, order sequences of opponents in multi-agent scenarios, etc.
    The ambition of this work is dual: 1) to present a compact and accessible introduction to the Automatic Curriculum Learning literature and 2) to draw a bigger picture of the current state of the art in ACL to encourage the cross-breeding of existing concepts and the emergence of new ideas.   
\end{abstract}

\section{Introduction}


Human learning is organized into a curriculum of interdependent learning situations of various complexities. For sure, Homer learned to formulate words before he could compose the Iliad. This idea was first transferred to machine learning in \mycite{selfridge}, where authors designed a \textit{learning scheme} to train a cart pole controller: first training on long and light poles, then gradually moving towards shorter and heavier poles. A related concept was also developed by \mycite{Schmid}, who proposed to improve world model learning by organizing exploration through \textit{artificial curiosity}.
In the following years, curriculum learning was applied to organize the presentation of training examples or the growth in model capacity in various supervised learning settings~\cite{elman,krueger,bengiocl}.
In parallel, the developmental robotics community proposed \textit{learning progress} as a way to self-organize open-ended \textit{developmental trajectories} of learning agents~\cite{oudeyer2007intrinsic}. Inspired by these earlier works, the Deep Reinforcement Learning (DRL) community developed a family of mechanisms called \textit{Automatic Curriculum Learning}, which we propose to define as follows:

\textit{\textbf{Automatic Curriculum Learning (ACL)} for DRL is a family of mechanisms that automatically adapt the distribution of training data by learning to adjust the selection of learning situations to the capabilities of DRL agents.}

\paragraph{Related fields.}ACL shares many connections with other fields. For example, ACL can be used in the context of \textit{Transfer Learning} where agents are trained on one distribution of tasks and tested on another~\cite{taylortransfer}. \textit{Continual Learning} trains agents to be robust to unforeseen changes in the environment while ACL assumes agents to stay in control of learning scenarios~\cite{continual-learning-review}. \textit{Policy Distillation} techniques \cite{pol-dil-review} form a complementary toolbox to target multi-task RL settings, where knowledge can be transferred from one policy to another (e.g. from task-expert policies to a generalist policy).

\paragraph{Scope.}This short survey proposes a typology of ACL mechanisms when combined with DRL algorithms and, as such, does not review population-based algorithms implementing ACL (e.g.~\mycite{imgep}, \mycite{poet}). 
As per our adopted definition, ACL refers to mechanisms \textit{explicitly} optimizing the automatic organization of training data. Hence, they should not be confounded with \textit{emergent curricula}, by-products of distinct mechanisms. For instance, the on-policy training of a DRL algorithm is not considered ACL, because the shift in the distribution of training data \textit{emerges} as a by-product of policy learning.
Given this is a short survey, we do not present the details of every particular mechanism. As the current ACL literature lacks theoretical foundations to ground proposed approaches in a formal framework, this survey focuses on empirical results.



\section{Automatic Curriculum Learning for DRL}
\label{sec:2}

This section formalizes the definition of ACL for Deep RL and proposes a classification. 

\paragraph{Deep Reinforcement Learning} \hspace{-0.2cm}is a family of algorithms which leverage deep neural networks for function approximation to tackle reinforcement learning problems. DRL agents learn to perform sequences of actions $a$ given states $s$ in an environment so as to maximize some notion of cumulative reward $r$~\cite{sutton2018reinforcement}.
Such problems are usually called \textit{tasks} and formalized as Markov Decision Processes (MDPs) of the form $T=\langle \mathcal{S},\mathcal{A},\mathcal{P}, \mathcal{R}, \rho_0 \rangle$ where $\mathcal{S}$ is the state space, $\mathcal{A}$ is the action space, $\mathcal{P}:S \times A \times S \rightarrow [0,1]$ is a transition function characterizing the probability of switching from the current state $s$ to the next state $s'$ given action $a$, $\mathcal{R}:S \times A  \rightarrow \mathbb{R}$ is a reward function and $\rho_0$ is a distribution of initial states. 
To challenge the generalization capacities of agents \cite{coinrun}, the community introduced multi-task DRL problems where agents are trained on tasks sampled from a task space: $T \sim \mathcal{T}$. In multi-goal DRL, policies and reward functions are conditioned on goals, which augments the task-MDP with a goal space $\mathcal{G}$~\cite{uvfa}.

\paragraph{Automatic Curriculum Learning} \hspace{-0.2cm} mechanisms propose to learn a task selection function $\mathcal{D}:\mathcal{\mathcal{H}\to\mathcal{T}}$ where $\mathcal{H}$ can contain any information about past interactions. This is done with the objective of maximizing a metric $P$ computed over a distribution of target tasks $\mathcal{T}_{target}$ after $N$ training steps:
\begin{equation}
    \label{eq:1}
    Obj: \max_{\mathcal{D}} \int_{T\sim \mathcal{T}_{target}} \! P_T^N\, \mathrm{d}T,
\end{equation}
where $P_T^N$ quantifies the agent's behavior on task $T$ after $N$ training steps (e.g. cumulative reward, exploration score). In that sense, ACL can be seen as a particular case of meta-learning, where $\mathcal{D}$ is learned along training to improve further learning. 

\paragraph{ACL Typology.} We propose a classification of ACL mechanisms based on three dimensions:
\begin{enumerate}[leftmargin=0.45cm, nolistsep]
    \item \textit{Why use ACL?} We review the different objectives that ACL has been used for (Section~\ref{sec:main_objective}).
    \item \textit{What does ACL control?} ACL can target different aspects of the learning problem (e.g. environments, goals, reward functions, Section~\ref{sec:lever})
    \item \textit{What does ACL optimize}? ACL mechanisms usually target surrogate objectives (e.g. learning progress, diversity) to alleviate the difficulty to optimize the main objective $Obj$ directly (Section~\ref{sec:surrogate_objective}).
\end{enumerate}


\section{Why use ACL?}
\label{sec:main_objective}
ACL mechanisms can be used for different purposes that can be seen as particular instantiations of the general objective defined in Eq~\ref{eq:1}.

\paragraph{Improving performance on a restricted task set.} Classical RL problems are about solving a given task, or a restricted task set (e.g. which vary by their initial state). In these simple settings, ACL has been used to improve sample efficiency or asymptotical performance~\cite{per,apex,SAUNA}.

\paragraph{Solving hard tasks.} Sometimes the target tasks cannot be solved directly (e.g. too hard or sparse rewards). 
In that case, ACL can be used to pose auxiliary tasks to the agent, gradually guiding its learning trajectory from simple to difficult tasks until the target tasks are solved. In recent works, ACL was used to schedule DRL agents from simple mazes to hard ones \cite{tscl}, or from close-to-success initial states to challenging ones in robotic control scenarios \cite{reverse-cur,BaRC} and video games \cite{montezuma-single-demo}. 
Another line of work proposes to use ACL to organize the exploration of the state space so as to solve sparse reward problems~\cite{countbased,icm,disagreement,pathakdisagreement,rnd}. In these works, the performance reward is augmented with an intrinsic reward guiding the agent towards uncertain areas of the state space.

\paragraph{Training generalist agents.} Generalist agents must be able to solve tasks they have not encountered during training (e.g. continuous task spaces or distinct training and testing set). ACL can shape learning trajectories to improve generalization, e.g. by avoiding unfeasible task subspaces \cite{portelas2019}. ACL can also help agents to generalize from simulation settings to the real world (Sim2Real)~\cite{OpenAI2019SolvingRC,ADRmila} or to maximize performance and robustness in multi-agent settings via Self-Play~\cite{alpha-go-zero,rarl,openaiSumos,Baker2019HidenSeek,vinyals2019grandmaster}.

\paragraph{Training multi-goal agents.}In multi-goal RL, agents are trained and tested on tasks that vary by their goals. Because agents can control the goals they target, they learn a behavioral repertoire through one or several goal-conditioned policies. The adoption of ACL in this setting can improve performance on a testing set of pre-defined goals. Recent works demonstrated the benefits of using ACL in scenarios such as multi-goal robotic arm manipulation \cite{her,eb-per,fournier-accuracy-acl,CGM,zhao2019curiosity,curriculu-her,curious} or multi-goal navigation \cite{asymetricSP,goalgan,settersolver,cideron2019self}.

\paragraph{Organizing open-ended exploration.}In some multi-goal settings, the space of achievable goals is not known in advance. Agents must discover achievable goals as they explore and learn how to represent and reach them. For this problem, ACL can be used to organize the discovery and acquisition of repertoires of robust and diverse behaviors, e.g. from visual observations~\cite{diayn,skewfit,metarl-carml} or from natural language interactions with social peers~\cite{le2,imagine}.

\section{What does ACL control?}
\label{sec:lever}

While \textit{on-policy} DRL algorithms directly use training data generated by the current behavioral policy, \textit{off-policy} algorithms can use trajectories collected from other sources. This practically decouples \textit{data collection} from \textit{data exploitation}. Hence, we organize this section into two categories: one reviewing ACL for data collection, the other ACL for data exploitation. 

\begin{figure}[h!]
\centering
\includegraphics[width=\columnwidth]{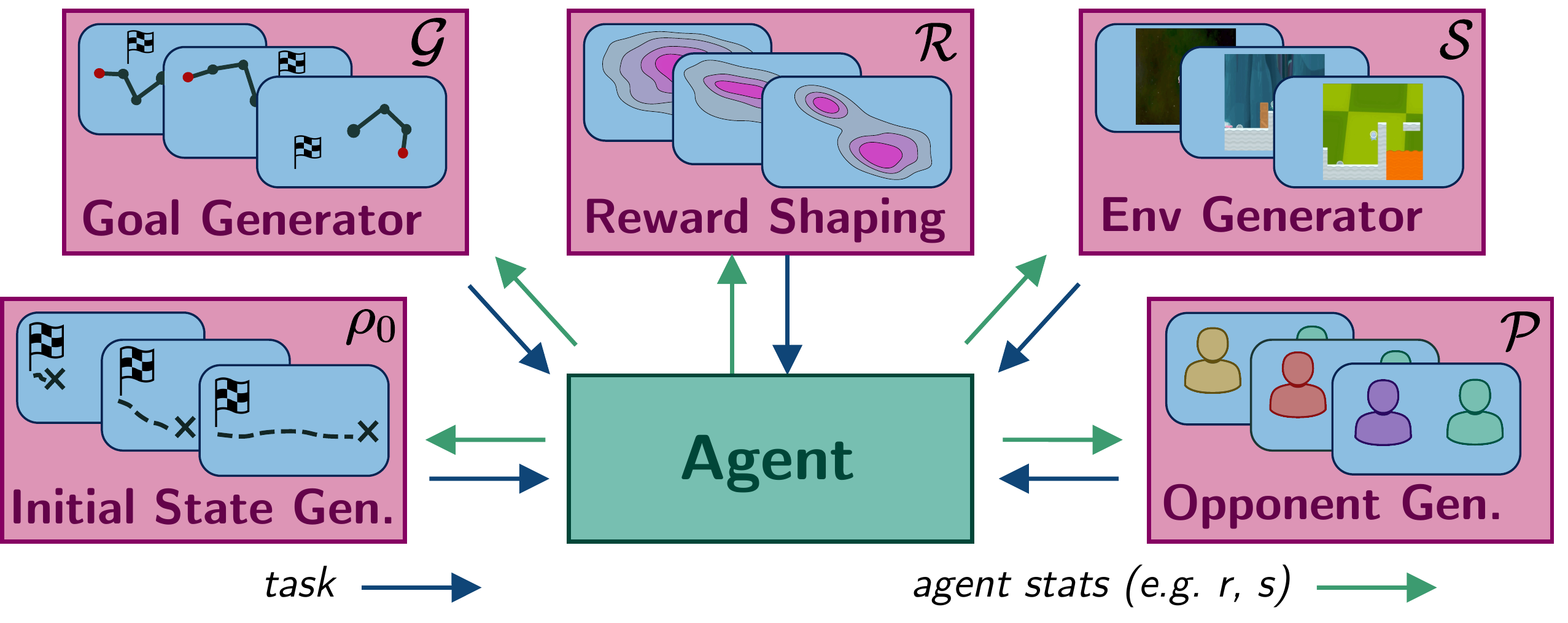}
\caption{ACL for data collection. ACL can control each elements of task MDPs to shape the learning trajectories of agents. Given metrics of the agent's behavior like performance or visited states, ACL methods generate new tasks adapted to the agent's abilities.}
\label{fancy-fig}
\end{figure}

\subsection{ACL for Data Collection}
\label{sec:lever_data_collection}


During data collection, ACL organizes the sequential presentation of tasks as a function of the agent's capabilities. To do so, it generates tasks by acting on elements of task MDPs (e.g. $\mathcal{R}, \mathcal{P}, \rho_0$, see Fig.~\ref{fancy-fig}). The curriculum can be designed on a discrete set of tasks or on a continuous task space. In single-task problems, ACL can define a set of auxiliary tasks to be used as stepping stones towards the resolution of the main task. The following paragraphs organize the literature according to the nature of the control exerted by ACL:

\paragraph{Initial state $(\rho_0)$.}The distribution of initial states $\rho_0$ can be controlled to modulate the difficulty of a task. Agents start learning from states close to a given target (i.e. easier tasks), then move towards harder tasks by gradually increasing the distance between the initial states and the target. This approach is especially effective to design auxiliary tasks for complex control scenarios with sparse rewards~\cite{reverse-cur,BaRC,montezuma-single-demo}.

\paragraph{Reward functions $(\mathcal{R})$.}ACL can be used for automatic reward shaping: adapting the reward function $\mathcal{R}$ as a function of the learning trajectory of the agent. In curiosity-based approaches especially, an internal reward function guides agents towards areas associated with high uncertainty to foster exploration~\cite{countbased,icm,disagreement,pathakdisagreement,rnd}. As the agent explores, uncertain areas --and thus the reward function-- change, which automatically devises a learning curriculum guiding the exploration of the state space. In~\mycite{fournier-accuracy-acl}, an ACL mechanism controls the tolerance in a goal reaching task. Starting with a low accuracy requirement, it gradually and automatically shifts towards stronger accuracy requirements as the agent progresses. In \mycite{diayn} and \mycite{metarl-carml}, authors propose to learn a skill space in unsupervised settings (from state space and pixels respectively), from which are derived reward functions promoting both behavioral diversity and skill separation.


\paragraph{Goals $(\mathcal{G})$.}In multi-goal DRL, ACL techniques can be applied to order the selection of goals from discrete sets~\cite{le2}, continuous goal spaces~\cite{asymetricSP,goalgan,skewfit,settersolver} or even sets of different goal spaces~\cite{CGM,curious}. Although goal spaces are usually pre-defined, recent work proposed to apply ACL on a goal space learned from \textit{pixels} using a generative model \cite{skewfit}. 

\paragraph{Environments $(\mathcal{S}, \mathcal{P})$.}ACL has been successfully applied to organize the selection of environments from a discrete set, e.g. to choose among Minecraft mazes~\cite{tscl} or Sonic the Hedgehog levels~\cite{tscllike}. A more general --and arguably more powerful-- approach is to leverage parametric \textit{Procedural Content Generation} (PCG) techniques~\cite{risiPCG} to generate rich task spaces. In that case, ACL allows to detect relevant niches of progress~\cite{OpenAI2019SolvingRC,portelas2019,ADRmila}.

\paragraph{Opponents~$(\mathcal{S}, \mathcal{P})$.}Self-play algorithms train agents against present or past versions of themselves~\cite{alpha-go-zero,openaiSumos,vinyals2019grandmaster,Baker2019HidenSeek}. The set of opponents directly maps to a set of tasks, as different opponents results in different transition functions $\mathcal{P}$ and possibly state spaces $\mathcal{S}$. Self-play can thus be seen as a form of ACL, where the sequence of opponents (i.e. tasks) is organized to maximize performance and robustness. In single-agent settings, an adversary policy can be trained to perturb the main agent~\cite{rarl}.

\subsection{ACL for Data Exploitation}
\label{sec:lever_data_exploitation}

ACL can also be used in the data exploitation stage, by acting on training data previously collected and stored in a \textit{replay memory}. It enables the agent to ``mentally experience the effects of its actions without actually executing them'', a technique known as \textit{experience replay}~\cite{lin1992self}. At the data exploitation level, ACL can exert two types of control on the distribution of training data: \textit{transition selection} and \textit{transition modification}.

\paragraph{Transition selection $(\mathcal{S}\times\mathcal{A})$.}Inspired from the \textit{prioritized sweeping} technique that organized the order of updates in planning methods~\cite{moore1993prioritized}, \mycite{per} introduced \textit{prioritized experience replay} (PER) for model-free off-policy RL to bias the selection of transitions for policy updates, as some transitions might be \textit{more informative} than others. Different ACL methods propose different metrics to evaluate the importance of each transition~\cite{per,eb-per,curious,zhao2019curiosity,le2,imagine}. Transition selection ACL techniques can also be used for on-policy algorithms to filter online learning batches \cite{SAUNA}.

\paragraph{Transition modification $(\mathcal{G})$.}In multi-goal settings, \textit{Hindsight Experience Replay} (HER) proposes to reinterpret trajectories collected with a given target goal with respect to a different goal \cite{her}. In practice, HER modifies transitions by substituting target goals $g$ with one of the outcomes $g'$ achieved later in the trajectory, as well as the corresponding reward $r'=R_{g'}(s,a)$. By explicitly biasing goal substitution to increase the probability of sampling rewarded transitions, HER shifts the training data distribution from simpler goals (achieved now) towards more complex goals as the agent makes progress. Substitute goal selection can be guided by other ACL mechanisms (e.g. favoring diversity~\cite{curriculu-her,cideron2019self}).


\section{What Does ACL Optimize?}
\label{sec:surrogate_objective}

 \begin{table*}[htb!]
\small
\centering
\begin{tabular}{llll}
\toprule
 \thead{Algorithm} & \thead{Why use ACL?} & \thead{What does ACL control?} & \thead{What does ACL optimize?} \\
 \midrule
ACL for Data Collection (\S~\ref{sec:lever_data_collection}):   &   &   &  \\
 \midrule
 
ADR (OpenAI) ~\cite{OpenAI2019SolvingRC}         &   Generalization              & Environments $(\mathcal{S},\mathcal{P})$ (PCG)    & Intermediate difficulty      \\
 
ADR (Mila) ~\cite{ADRmila}                       &   Generalization              & Environments $(\mathcal{P})$ (PCG)           & Intermediate diff. \& Diversity               \\
ALP-GMM ~\cite{portelas2019}                     & Generalization                & Environments $(\mathcal{S})$ (PCG)           & LP                       \\

  RARL ~\cite{rarl}                                &  Generalization      & Opponents $(\mathcal{P})$          & ARM                     \\
 AlphaGO Zero ~\cite{alpha-go-zero}          &  Generalization                       & Opponents $(\mathcal{P})$          & ARM                     \\
 Hide\&Seek ~\cite{Baker2019HidenSeek}            &  Generalization                       & Opponents $(\mathcal{P})$           & ARM                \\
 AlphaStar ~\cite{vinyals2019grandmaster}       &  Generalization                       & Opponents $(\mathcal{P})$          & ARM \& Diversity               \\
 Competitive SP ~\cite{openaiSumos}                &  Generalization                       & Opponents $(\mathcal{P})$           & ARM \& Diversity               \\
  RgC ~\cite{tscllike}                             & Generalization                & Environments $(\mathcal{S})$ (DS)            & LP                      \\
 
RC ~\cite{reverse-cur}                &  Hard Task   & Initial states $(\rho_0)$         & Intermediate difficulty                  \\
$1$-demo RC~\cite{montezuma-single-demo}               &   Hard Task              & Initial states $(\rho_0)$              & Intermediate difficulty                  \\
 Count-based ~\cite{countbased}                   &  Hard Task                   & Reward functions $(\mathcal{R})$                  & Diversity              \\
  RND ~\cite{rnd}                                 &  Hard Task                   & Reward functions $(\mathcal{R})$                  & Surprise (model error)              \\
 ICM ~\cite{icm}                                  &  Hard Task                   & Reward functions $(\mathcal{R})$                  & Surprise (model error)              \\
 Disagreement ~\cite{pathakdisagreement}          &  Hard Task                   & Reward functions  $(\mathcal{R})$                 & Surprise (model disagreement)              \\
 MAX ~\cite{disagreement}                &  Hard Task                   & Reward functions  $(\mathcal{R})$                & Surprise (model disagreement)              \\

BaRC ~\cite{BaRC}                                &   Hard Task              & Initial states $(\rho_0)$              & Intermediate difficulty                  \\

 TSCL ~\cite{tscl}                                & Hard Task                 & Environments $(\mathcal{S})$ (DS)            & LP                       \\
 
  Acc-based CL ~\cite{fournier-accuracy-acl}       & Multi-Goal        & Reward function $(\mathcal{R})$            & LP                      \\

Asym. SP ~\cite{asymetricSP}                     &  Multi-Goal     & Goals $(\mathcal{G})$, initial states $(\rho_0)$   & Intermediate difficulty   \\
   
GoalGAN ~\cite{goalgan}                          &   Multi-Goal      & Goals  $(\mathcal{G})$            & Intermediate difficulty             \\
 
Setter-Solver ~\cite{settersolver}               &   Multi-Goal               & Goals  $(\mathcal{G})$                      & Intermediate difficulty                    \\

CGM ~\cite{CGM}                          & Multi-Goal & Goals $(\mathcal{G})$           & Intermediate difficulty      \\

CURIOUS ~\cite{curious}                          & Multi-Goal & Goals $(\mathcal{G})$           & LP      \\

 Skew-fit ~\cite{skewfit}                         & Open-Ended Explo.       & Goals $(\mathcal{G})$ (from pixels)          & Diversity                    \\
  DIAYN \cite{diayn} & Open-Ended Explo. & Reward functions $(\mathcal{R})$ & Diversity \\
 CARML ~\cite{metarl-carml}             & Open-Ended Explo.                        & Reward functions $(\mathcal{R})$          & Diversity       \\

 LE2 ~\cite{le2}                                  & Open-Ended Explo. & Goals $(\mathcal{G})$             & Reward \& Diversity      \\
 \midrule
 ACL for Data Exploitation  (\S~\ref{sec:lever_data_exploitation}):    &               &                  &                          \\
 \midrule
  Prioritized ER ~\cite{per}                       & Performance boost    & Transition selection $(\mathcal{S}\times\mathcal{A})$    & Surprise (TD-error)                 \\
  SAUNA ~\cite{SAUNA}                              & Performance boost    & Transition selection $(\mathcal{S}\times\mathcal{A})$    & Surprise (V-error)               \\ 
  CURIOUS ~\cite{curious}                          &   Multi-goal         & Trans. select. \& mod. $(\mathcal{S}\times\mathcal{A}, \mathcal{G})$          & LP \& Energy      \\
 HER ~\cite{her}      & Multi-goal   & Transition modification $(\mathcal{G})$   & Reward             \\
 
HER-curriculum ~\cite{curriculu-her}     & Multi-goal  & Transition modification $(\mathcal{G})$ &     Diversity                  \\
 
Language HER ~\cite{cideron2019self}              & Multi-goal        & Transition modification   $(\mathcal{G})$           & Reward              \\
 
Curiosity Prio. ~\cite{zhao2019curiosity} & Multi-goal  & Transition selection $(\mathcal{S}\times\mathcal{A})$            & Diversity         \\
 
En. Based ER ~\cite{eb-per}      &  Multi-goal            & Transition selection  $(\mathcal{S}\times\mathcal{A})$   & Energy                \\

 LE2 ~\cite{le2}                                  & Open-Ended Explo.            & Trans. select. \& mod. $(\mathcal{S}\times\mathcal{A}, \mathcal{G})$          & Reward      \\
  IMAGINE ~\cite{imagine}                         & Open-Ended Explo.            & Trans. select. \& mod. $(\mathcal{S}\times\mathcal{A}, \mathcal{G})$          & Reward      \\

 \bottomrule
\end{tabular}
\caption{Classification of the surveyed papers. The classification is organized along the three dimensions defined in the above text. In \textit{Why use ACL}, we only report the main objective of each work. When ACL controls the selection of environments, we precise whether it is selecting them from a discrete set (\textit{DS}) or through parametric Procedural Content Generation (\textit{PCG}). We abbreviate \textit{adversarial reward maximization} by \textit{ARM} and \textit{learning progress} by \textit{LP}.}
\label{bigtable}
\end{table*}

Objectives such as the average performance on a set of testing tasks after $N$ training steps can be difficult to optimize directly. To alleviate this difficulty, ACL methods use a variety of surrogate objectives.

\paragraph{Reward.}As DRL algorithms learn from reward signals, rewarded transitions are usually considered as more informative than others, especially in sparse reward problems.
In such problems, ACL methods that act on transition selection may artificially increase the ratio of high versus low rewards in the batches of transitions used for policy updates~\cite{narasimhan2015language,unreal,imagine}.
In multi-goal RL settings where some goals might be much harder than others, this strategy can be used to balance the proportion of positive rewards for each of the goals~\cite{curious,le2}.
Transition modification methods favor rewards as well, substituting goals to increase the probability of observing rewarded transitions~\cite{her,cideron2019self,le2,imagine}.
In data collection however, adapting training distributions towards more rewarded experience leads the agent to focus on tasks that are already solved. Because collecting data from already solved tasks hinders learning, data collection ACL methods rather focus on other surrogate objectives.

\paragraph{Intermediate difficulty.} A more natural surrogate objective for data collection is \textit{intermediate difficulty}. Intuitively, agents should target tasks that are neither too easy (already solved) nor too difficult (unsolvable) to maximize their learning progress. Intermediate difficulty has been used to adapt the distribution of initial states from which to perform a hard task \cite{reverse-cur,montezuma-single-demo,BaRC}. This objective is also implemented in {GoalGAN}, where a curriculum generator based on a Generative Adversarial Network is trained to propose goals for which the agent reaches intermediate performance~\cite{goalgan}.
\mycite{settersolver} further introduced a \textit{judge network} trained to predict the feasibility of a given goal for the current learner. Instead of labelling tasks with an intermediate level of difficulty as in GoalGAN, this Setter-Solver model generates goals associated to a random feasibility uniformly sampled from $[0,1]$. The type of goals varies as the agent progresses, but the agent is always asked to perform goals sampled from a distribution balanced in terms of feasibility. 
In \mycite{asymetricSP}, tasks are generated by an RL policy trained to propose either goals or initial states so that the resulting navigation task is of intermediate difficulty w.r.t. the current agent. 
Intermediate difficulty ACL has also been driving successes in Sim2Real applications, where it sequences \textit{domain randomizations} to train policies that are robust enough to generalize from simulators to real-world robots~\cite{ADRmila,OpenAI2019SolvingRC}. \mycite{OpenAI2019SolvingRC} trains a robotic hand control policy to solve a Rubik's cube by automatically adjusting the task distribution so that the agent achieves decent performance while still being challenged.

\paragraph{Learning progress.}The $Obj$ objective of ACL methods can be seen as the  maximization of a \textit{global learning progress}: the difference between the final score $\int_{T\sim \mathcal{T}} \! P_T^N\, \mathrm{d}T$ and the initial score $\int_{T\sim \mathcal{T}} \! P_T^0\, \mathrm{d}T$. To approximate this complex objective, measures of competence learning progress (LP) localized in space and time were proposed in earlier developmental robotics works~\cite{baranes2013active,imgep}. Like \textit{Intermediate difficulty}, maximizing LP drives learners to practice tasks that are neither too easy nor too difficult, but LP does not require a threshold to define what is "intermediate" and is robust to tasks with intermediate scores but where the agent cannot improve. 
LP maximization is usually framed as a multi-armed bandit (MAB) problem where tasks are arms and their LP measures are associated values. Maximizing LP values was shown optimal under the assumption of concave learning profiles~\cite{lopes2012strategic}. Both \mycite{tscl} and \mycite{tscllike} measure LP as the estimated derivative of the performance for each task in a discrete set (Minecraft mazes and Sonic the Hedgehog levels respectively) and apply a MAB algorithm to automatically build a curriculum for their learning agents. At a higher level, CURIOUS uses \textit{absolute} LP to select goal \textit{spaces} to sample from in a simulated robotic arm setup~\cite{curious} (absolute LP enables to redirect learning towards tasks that were forgotten or that changed). There, absolute LP is also used to bias the sampling of transition used for policy updates towards high-LP goals. ALP-GMM uses absolute LP to organize the presentation of procedurally-generated Bipedal-Walker environments sampled from a continuous task space through a stochastic parameterization~\cite{portelas2019}. They leverage a Gaussian Mixture Model to recover a MAB setup over the continuous task space. LP can also be used to guide the choice of accuracy requirements in a reaching task~\cite{fournier-accuracy-acl}, or to train a \textit{replay policy} via RL to sample transitions for policy updates~\cite{exp-replay-opti}.

\paragraph{Diversity.}Some ACL methods choose to maximize measures of diversity (also called novelty or low density). 
In multi-goal settings for example, ACL might favor goals from low-density areas either as targets~\cite{skewfit} or as substitute goals for data exploitation~\cite{curriculu-her}. Similarly, \mycite{zhao2019curiosity} biases the sampling of trajectories falling into low density areas of the trajectory space. %
In single-task RL, \textit{count-based} approaches introduce internal reward functions as decreasing functions of the state visitation count, guiding agent towards rarely visited areas of the state space~\cite{countbased}.
Through a variational expectation-maximization framework, \mycite{metarl-carml} propose to alternatively update a latent skill representation from experimental data (as in \mycite{diayn}) and to meta-learn a policy to adapt quickly to tasks constructed by deriving a reward function from sampled skills.
Other algorithms do not optimize directly for diversity but use heuristics to maintain it. For instance, \mycite{portelas2019} maintains exploration by using a residual uniform task sampling and \mycite{openaiSumos} sample opponents from past versions of different policies to maintain diversity. 

\paragraph{Surprise.} Some ACL methods train transition models and compute intrinsic rewards based on their prediction errors~\cite{icm,rnd} or based on the disagreement (variance) between several models from an ensemble~\cite{disagreement,pathakdisagreement}. The general idea is that models tend to give bad prediction (or disagree) for states rarely visited, thus inducing a bias towards less visited states. However, a model might show high prediction errors on stochastic parts of the environment (TV problem~\cite{icm}), a phenomenon that does not appear with model disagreement, as all models of the ensemble eventually learn to predict the (same) mean prediction \cite{pathakdisagreement}. 
Other works bias the sampling of transitions for policy update depending on their temporal-difference error (TD-error), i.e. the difference between the transition's value and its next-step bootstrap estimation~\cite{per,apex}. Similarly, \mycite{SAUNA} adapt transition selection based on the discrepancy between the observed return and the prediction of the value function of a PPO learner (V-error). Whether the error computation involves value models or transition models, ACL mechanisms favor states related to maximal~\textit{surprise}, i.e. a maximal difference between the expected (model prediction) and the truth.

\paragraph{Energy.} In the data exploitation phase of multi-goal settings, \mycite{eb-per} prioritize transitions from \textit{high-energy} trajectories (e.g. kinetic energy) while \mycite{curious} prioritize transitions where the object relevant to the goal moved (e.g. cube movement in a cube pushing task).

\paragraph{Adversarial reward maximization (ARM).} Self-Play is a form of ACL which optimizes agents' performance when opposed to current or past versions of themselves, an objective that we call \textit{adversarial reward maximization (ARM)} \cite{self-play-framework}. While agents from \mycite{alpha-go-zero} and \mycite{Baker2019HidenSeek} always oppose copies of themselves, \mycite{openaiSumos} train several policies in parallel and fill a pool of opponents made of current and past versions of all policies. This maintains a diversity of opponents, which helps to fight catastrophic forgetting and to improve robustness.
%
In the multi-agent game Starcraft~II, \mycite{vinyals2019grandmaster} train three main policies in parallel (one for each of the available player types). They maintain a \textit{league} of opponents composed of current and past versions of both the three main policies and additional adversary policies. Opponents are not selected at random but to be challenging (as measured by winning rates).
 

\section{Discussion}

\paragraph{The bigger picture.}In this survey, we unify the wide range of ACL mechanisms used in symbiosis with DRL under a common framework. ACL mechanisms are used with a particular goal in mind (e.g. organizing exploration, solving hard tasks, etc. \S~\ref{sec:main_objective}). It controls a particular element of task MDPs (e.g. $\mathcal{S}, \mathcal{R}, \rho_0$, \S~\ref{sec:lever}) and maximizes a surrogate objective to achieve its goal (e.g. diversity, learning progress, \S~\ref{sec:surrogate_objective}). Table~\ref{bigtable} organizes the main works surveyed here along these three dimensions. Both previous sections and Table~\ref{bigtable} present what has been implemented in the past, and thus, by contrast, highlight potential new avenues for ACL.


\noindent\textit{Expanding the set of ACL targets.} Inspired by the maturational mechanisms at play in human infants, \mycite{elman} proposed to gradually expand the working memory of a recurrent model in a word-to-word natural language processing task. The idea of changing the properties of the agent (here its memory) was also studied in developmental robotics~\cite{mature}, policy distillation methods~\cite{mixmatch,pol-dil-review} and evolutionary approaches \cite{ha-design}  but is absent from the ACL-DRL literature. ACL mechanisms could indeed be used to control the agent's body ($\mathcal{S}, \mathcal{P}$), its action space (how it acts in the world, $\mathcal{A}$), its observation space (how it perceives the world, $\mathcal{S}$), its learning capacities (e.g. capacities of the memory, or the controller) or the way it perceives time (controlling discount factors~\cite{cl-discount}).

\noindent\textit{Combining approaches.} Many combinations of previously defined ACL mechanisms remain to be investigated. Could we use LP to optimize the selection of opponents in self-play approaches? To drive goal selection in learned goal spaces (e.g. \mycite{Finot2019}, population-based)? Could we train an adversarial domain generator to robustify policies trained for Sim2Real applications?


\paragraph{On the need of systematic ACL studies.}Given the positive impact that ACL mechanisms can have in complex learning scenarios, one can only deplore the lack of comparative studies and standard benchmark environments. Besides, although empirical results advocate for their use, a theoretical understanding of ACL mechanisms is still missing. Although there have been attempts to frame CL in supervised settings \cite{bengiocl,hacohen19a-scoring-pacing}, more work is needed to see whether such considerations hold in DRL scenarios.

\paragraph{ACL as a step towards open-ended learning agents.}\hspace{-0.2cm}Alan Turing famously wrote \textit{``Instead of trying to produce a programme to simulate the adult mind, why not rather try to produce one which simulates the child's?''} \cite{turing1950computing}. The idea of starting with a simple machine and to enable it to learn autonomously is the cornerstone of developmental robotics but is rarely considered in DRL \cite{imagine,diayn,metarl-carml}. Because they actively organize learning trajectories as a function of the agent's properties, ACL mechanisms could prove extremely useful in this quest. We could imagine a learning architecture leveraging ACL mechanisms to control many aspects of the learning odyssey, guiding agents from their simple original state towards fully capable agents able to reach a multiplicity of goals. As we saw in this survey, these ACL mechanisms could control the development of the agent's body and capabilities (motor actions, sensory apparatus), organize the exploratory behavior towards tasks where agents learn the most (maximization of information gain, competence progress) or guide acquisitions of behavioral repertoires.

\scalefont{0.95}
\bibliographystyle{named}
\bibliography{ijcai20}

\end{document}